%% file: example_paper.tex
\documentclass{article}

\usepackage{microtype}
\usepackage{graphicx}
\usepackage{subfigure}
\usepackage{booktabs} 
\usepackage{xspace}
\usepackage{multirow}
\usepackage{tabularx}
\usepackage{makecell}
\usepackage{bm} 
\usepackage{xcolor} 
\usepackage{colortbl}
\usepackage{enumitem}
\usepackage{float}
\usepackage{diagbox}

\usepackage[colorlinks,
linkcolor=blue,
anchorcolor=blue,
citecolor=purple,
urlcolor=purple,
backref=page]{hyperref}

\newcommand{\model}{LBP\xspace}
\newcolumntype{x}[1]{>{\centering\arraybackslash}p{#1pt}}

\usepackage[accepted]{icml2025}

\usepackage{amsmath}
\usepackage{amssymb}
\usepackage{mathtools}
\usepackage{amsthm}
\usepackage[capitalize,noabbrev]{cleveref}

\theoremstyle{plain}

\theoremstyle{definition}

\theoremstyle{remark}

\usepackage[textsize=tiny]{todonotes}

\icmltitlerunning{Efficient Robotic Policy Learning via Latent Space Backward Planning}

\begin{document}

\twocolumn[
\icmltitle{Efficient Robotic Policy Learning via Latent Space Backward Planning}



\icmlsetsymbol{equal}{*}
\icmlsetsymbol{corr}{$\dagger$}

\begin{icmlauthorlist}
\icmlauthor{Dongxiu Liu}{thu,bupt,equal}
\icmlauthor{Haoyi Niu}{thu,equal}
\icmlauthor{Zhihao Wang}{thu,pku}
\icmlauthor{Jinliang Zheng}{thu,ailab}
\icmlauthor{Yinan Zheng}{thu}
\icmlauthor{Zhonghong Ou}{bupt,corr}
\icmlauthor{Jianming Hu}{thu,corr}
\icmlauthor{Jianxiong Li}{thu}
\icmlauthor{Xianyuan Zhan}{thu,ailab,corr}

\end{icmlauthorlist}

\icmlaffiliation{thu}{Tsinghua University}
\icmlaffiliation{pku}{Peking University}
\icmlaffiliation{bupt}{Beijing University of Posts and Telecommunications}
\icmlaffiliation{ailab}{Shanghai AI Lab}

\icmlcorrespondingauthor{Zhonghong Ou}{zhonghong.ou@bupt.edu.cn}
\icmlcorrespondingauthor{Jianming Hu}{hujm@mail.tsinghua.edu.cn}
\icmlcorrespondingauthor{Xianyuan Zhan}{zhanxianyuan@air.tsinghua.edu.cn}

\icmlkeywords{Machine Learning, ICML}

\vskip 0.3in
]



\printAffiliationsAndNotice{\icmlEqualContribution. Work done during internship at Institute for AI Industry Research (AIR), Tsinghua University.} 

\begin{abstract}
Current robotic planning methods often rely on predicting multi-frame images with full pixel details. While this fine-grained approach can serve as a generic world model, it introduces two significant challenges for downstream policy learning: substantial computational costs that hinder real-time deployment, and accumulated inaccuracies that can mislead action extraction. Planning with coarse-grained subgoals partially alleviates efficiency issues. However, their forward planning schemes can still result in off-task predictions due to accumulation errors, leading to misalignment with long-term goals. This raises a critical question: Can robotic planning be both efficient and accurate enough for real-time control in long-horizon, multi-stage tasks?
To address this, we propose a \textbf{L}atent Space \textbf{B}ackward \textbf{P}lanning scheme (\textbf{LBP}), which begins by grounding the task into final latent goals, followed by recursively predicting intermediate subgoals closer to the current state. The grounded final goal enables backward subgoal planning to always remain aware of task completion, facilitating on-task prediction along the entire planning horizon. The subgoal-conditioned policy incorporates a learnable token to summarize the subgoal sequences and determines how each subgoal guides action extraction.
Through extensive simulation and real-robot long-horizon experiments, we show that LBP outperforms existing fine-grained and forward planning methods, achieving SOTA performance. Project Page: \url{https://lbp-authors.github.io}
\end{abstract}

\section{Introduction}\label{intro}
Accurately predicting future states is crucial for many robotic planning methods in solving long-horizon, multi-stage tasks, where models must anticipate outcomes over extended temporal sequences. However, this requires balancing two conflicting objectives: (1) capturing sufficiently rich and accurate future information for task completion, and (2) maintaining computational efficiency for real-time decision making. Current methods often struggle with this trade-off. Those that prioritize long-term capability typically predict multi-step future states to provide detailed guidance, but are prone to high computational costs and rapid error accumulation. In contrast, efficiency-oriented methods often sacrifice the semantic richness needed to handle complex, long-horizon tasks.
This creates a fundamental trilemma—balancing efficiency, accuracy and sufficient future guidance—that remains unresolved and presents a significant challenge in real-world robotic planning.

To model future outcomes, one category of existing robotic planning methods~\citep{du2024learning_Unipi,ajay2024compositional,hu2024video_VPP} resorts to predicting an episode of future video as policy guidance.
However, predicting consecutive frames can propagate inaccuracies that compound over time, resulting in significant deviations from the intended final goal or physically inconsistent frames that confuse the downstream policies.
Furthermore, modeling entire future videos requires high computational costs and puts a heavy burden on real-time inference.
Obviously, predicting every detail in the future is often unnecessary for task execution, and the fine-grained predictions hinder both computational efficiency and task-oriented consistency.

The second category of robotic planning methods focuses on predicting future subgoals~\citep{Nair2020Hierarchical_HVF,huang2024subgoal}. These coarse-grained subgoals improve planning efficiency and reduce computational burden.
However, it still adheres to the forward planning paradigm, often leading to plans that are less aligned with distant goals, which in turn cause off-task behavior~\citep{kang2024incorporating_TaKSIE}. To address this, recent methods have introduced reachability or optimality checks~\citep{eysenbach2019search,nasiriany2019planning,fang2022planning,huang2024subgoal} to correct deviations and improve on-task accuracy. However, these post-hoc adjustments also add lots of complexities and do not really address the fundamental challenges.

The two aforementioned categories of methods both have some pros and cons. 
The video planning methods provide rich future guidance but suffer from heavy training demands and inefficient inference. The subgoal planning methods enjoy efficient planning but trade off long-horizon task progress guidance. 
Apart from failing to strike a desirable balance across different considerations, all previous efforts fall short in maintaining on-task prediction accuracy.
How can we address the above limitations and enable robots to plan efficiently and effectively through long-horizon tasks? 

\begin{figure}[t]
	\centering	\includegraphics[width=.5\textwidth]{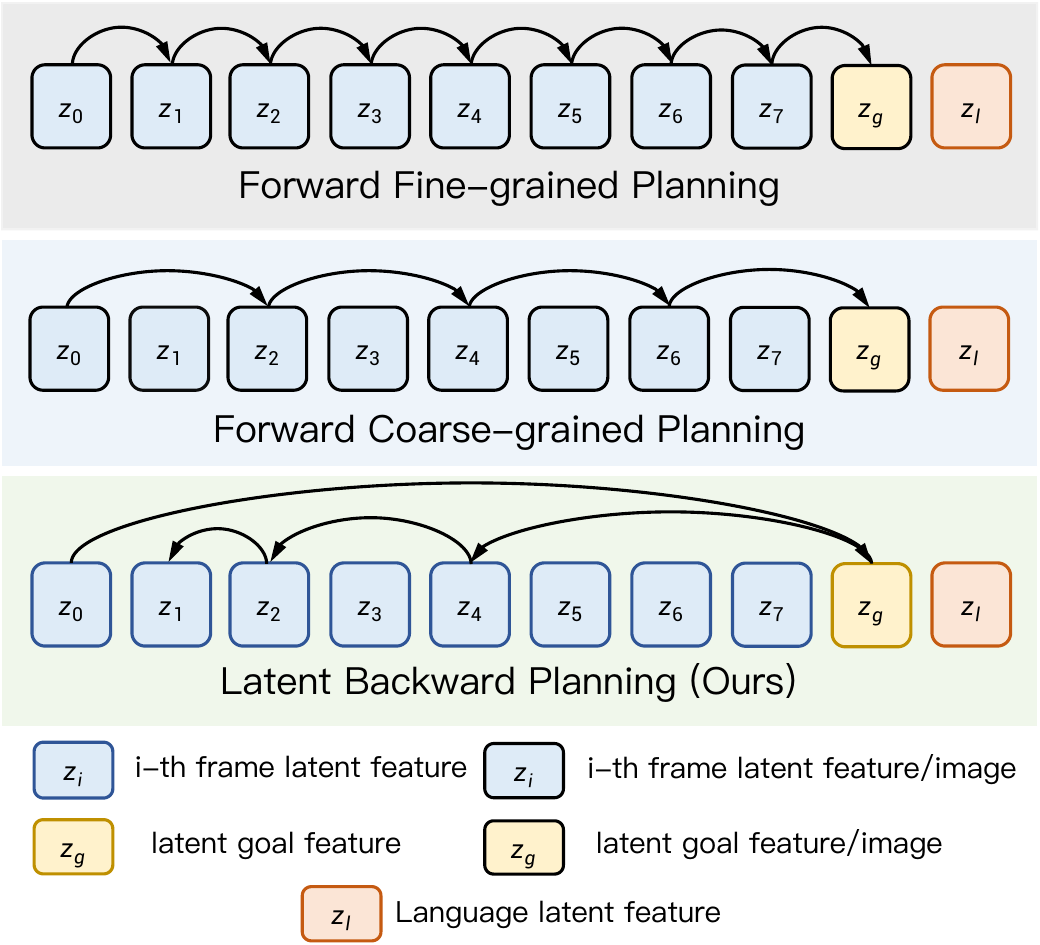}
        \vspace{-0.75cm}
		\caption{Illustration of latent space backward planning.}
        
        \label{fig:intro}
        \vspace{-0.5cm}
	\end{figure}

In this paper, we propose a \textbf{B}ackward \textbf{P}lanning approach in \textbf{L}atent space (\textbf{LBP}) for language-guided robotic control as in Figure~\ref{fig:intro}.
LBP first trains a latent goal predictor that maps the current state and language description to a distant final goal, grounding the task objective in latent image space to enforce task progression.
Second, LBP recursively predicts intermediate subgoals that are closer to the current state, ensuring that each subgoal remains aligned with the task progression and completion.
These two steps mirror how humans plan in complex tasks: we begin by envisioning the desired outcome based on the task objective, and then break it down into smaller, gradually manageable subgoals that are closer to the present stage. 
The subgoal sequences in LBP track the path to the goal with less redundancy, providing denser guidance in near terms while preserving task progression information over the entire planning horizon.
Lastly, LBP incorporates a subgoal fusion technique that enables the subgoal-conditioned policy to adaptively determine how to best utilize subgoals at varying distances. 
Collectively, LBP effectively addresses the triplet of challenges of off-task planning, insufficient guidance, and high computational costs inherent in previous methods.

LBP provides
a lightweight planning framework for robotic policy learning with on-task subgoal generation guarantee.
It combines the strengths of latent planning~\citep{wang2023mimicplay,wen2023any_ATM} and coarse-grained subgoal planning~\citep{wang-niu2024rsp,blackzero_SuSIE}, drastically reducing computational costs and enabling real-time deployment. 
Unlike previous methods that struggle with subgoal horizon selection, LBP provides an informative subgoal sequence spanning the entire planning horizon toward the final goal. 
This offers flexibility that allows the downstream policy to leverage subgoal signals at varying distances. 
The backward planning approach further enhances on-task accuracy by ensuring that the predicted subgoal sequence remains aligned with the overall task progression and the ultimate objective.
Through extensive evaluations in both simulation and real-robot experiments, we demonstrate that LBP significantly outperforms existing methods, especially excelling on long-horizon, multi-stage tasks.

\section{Related Works}
\paragraph{Video Planning.}
A significant body of research has explored video generation as planners for visuomotor control~\citep{,pertsch2020long_gcp,du2024learning_Unipi,ajay2024compositional,hu2024video_VPP,wu2024unleashing,bharadhwaj2024gen2act}. Approaches such as UniPi~\citep{du2024learning_Unipi} and HiP~\citep{ajay2024compositional} generate actions using inverse dynamics models from predicted consecutive frames, while Seer~\citep{tian2024predictive_SEER} and GR-1~\citep{wu2024unleashing} jointly predict actions and subsequent image frames. Although some methods operate in latent space~\citep{nair2020goal,hu2024video_VPP,li2024decisionnce}, this line of work faces significant challenges, including high computational demands and limited real-time capabilities, primarily due to the need to generate every consecutive frame of the future. Most of these methods operate in a forward autoregressive manner~\citep{wu2024unleashing,tian2024predictive_SEER}, which are prone to rapid error accumulation over time, significantly complicating policy learning.
In summary, these approaches attempt to plan with excessive detail that is often unnecessary to visuomotor control, resulting in computational inefficiency, compounded prediction errors, and challenges in effective action extraction.

\paragraph{Coarse-grained Planning.}
Coarse-grained planning approaches focus on predicting intermediate subgoals~\citep{Nair2020Hierarchical_HVF,wang2023mimicplay,blackzero_SuSIE,wang-niu2024rsp}, improving computational efficiency by avoiding the need to predict every frame of details. Goal-conditioned supervised (GCSL)~\citep{ghosh2021learning,emmons2022rvs,wang-niu2024rsp} and reinforcement learning (GCRL)~\citep{chane-sane2021RIS,por, park2024hiql}, have demonstrated that planning with intermediate goals can alleviate downstream policy learning burdens while enhancing long-horizon capabilities in simulation benchmark tasks. However, this paradigm faces unresolved challenges, particularly in subgoal selection, such as 1) determining appropriate prediction horizons, and 2) balancing the number of subgoals for effective policy guidance~\citep{levy2018hierarchical,nachum2020why}. Distant subgoals provide limited actionable information, while nearby subgoals may misalign with final task objectives. Similarly, excessive subgoals increase model complexity, whereas sparse subgoals fail to capture task progression semantics. Existing methods lack principled treatment to balance planning efficiency and long-term reliability.
Furthermore, the forward planning paradigm inherently suffers from error accumulation over time, leading to off-task behavior~\citep{kang2024incorporating_TaKSIE}. Recent attempts to mitigate this through post-hoc corrections, such as reachability or optimality checks~\citep{eysenbach2019search,nasiriany2019planning,fang2022planning,huang2024subgoal}, which add extra complexity without addressing the fundamental limitations of forward planning.

\vspace{-4pt}
\paragraph{Summary.}
Both fine-grained (video) and coarse-grained (subgoal) planning approaches fail to resolve the fundamental trilemma of robotic planning: achieving computational efficiency, maintaining long-horizon consistency, and ensuring prediction accuracy. These limitations highlight the need for a novel approach that appropriately balances these objectives in long-horizon, multi-stage visuomotor tasks. Inspired by the ``coarse-to-fine'' paradigm in natural language processing~\citep{wei2022chain},  computer vision~\citep{tian2024visual}, and decision making~\cite{wang-niu2024rsp}, we propose a backward planning framework that predicts subgoals in reverse temporal order—from coarse to fine horizons—starting from the final goal. At every control step, this approach generates a subgoal sequence that spans the entire task horizon, providing sufficient actionable guidance efficiently while minimizing error accumulation by ensuring consistent task alignment.

\vspace{-2pt}
\section{Preliminaries}
We consider the problem of learning a visuomotor policy conditioned on different contexts \( c \) that reflect task objective or completeness. These contexts $c$ can include goal images \( I_g \in \mathcal{G} \subset \mathcal{I} \), language descriptions \( l \in \mathcal{L} \), intermediate subgoals \( w_i \in \mathcal{W} \subset \mathcal{I} \), and etc.
Each video segment is defined as \( \tau_i = \{(I_t, a_t)\}^{H_i} \) with $H_i$ frames, where \( I_t \in \mathcal{I} \) represents the image observation and \( a_t \in \mathcal{A} \) denotes the action at time step \( t \). Given a dataset of video segments \( \mathcal{D} = \{\tau_1, \tau_2, \dots, \tau_N\} \) and a distribution over contexts \( f(c|\tau) \), a conditioned policy \( \pi_\theta(a|I, c) \) is trained to generate control signals in a closed-loop manner, achieving the desired behaviors aligned with the task description or future goals. The policy learning objective can be given as:
\begin{equation}
    \max_\theta \sum_{\tau \in \mathcal{D}} \sum_{1 \leq t \leq H} \mathbb{E}_{c \sim f(c|\tau)}[\log \pi_\theta(a_t|I_t, c)]
\end{equation}

We use the expectation over all contexts because some video segments are annotated with multiple types of task-relevant information, which can be utilized as guidance during policy learning. For instance, \( f(l|\tau) \) represents the distribution of language descriptions, \( f(I_g, l|\tau) = f(I_g|I_t,l) f(l|\tau) \) models the joint distribution of goal images and language descriptions, and \( f(w, I_g, l|\tau) = f(w = I_{t+k}|I_t, I_g, l) f(I_g|I_t,l) f(l|\tau) \) captures the distribution of \( k \)-step future subgoals, goal images, and language descriptions. Each context provides a different level of guidance: language serves as a basic task identifier, goal images indicate task completeness, and subgoals reflect task progression toward completion. By exploring different combinations of these contexts, we can adapt the level of guidance to meet varying demands for granularity in policy learning.


\section{Latent Backward Planning}
\paragraph{Overview.} We propose latent space backward planning (LBP), an efficient and robust planning framework for long-horizon visuomotor tasks, built upon the idea of backward subgoal prediction.
We observe that existing long-horizon planning with predicted subgoals struggles with (1) planning inefficiency and (2) off-task prediction.
Generating high-dimensional subgoal images poses significant challenges of computation loads, while modeling every future frame sequentially further deteriorates temporal efficiency, collectively hindering real-time real-world planning.

Thus, one of our key insights is planning in latent space with coarse-grained subgoals, enhancing planning efficiency in both spatial and temporal dimensions.
While latent subgoal planning has been explored in existing works~\citep{veerapaneni2020entity}, they typically adopt forward planning that often fail to align subgoals with ultimate task objectives.
Without accounting for task completion, subgoals can easily deviate from desired task progression, causing downstream policies to suffer from compounding errors snowballing along the planning process.
Existing approaches have to introduce additional subgoal quality checks on reachability or optimality to combat the error accumulation~\citep{Nair2020Hierarchical_HVF,eysenbach2019search,nasiriany2019planning,fang2022planning,huang2024subgoal}, at the cost of adding unnecessary complexity and trading off efficiency, but without fundamentally resolving the underlying off-task issues.

Another key insight of ours is that we can learn a final goal predictor that grounds the ultimate task objective (i.e. language description) into latent image space (Section~\ref{ground}).
Latent image space encapsulates much richer task progression information than language space, enabling backward planned subgoal sequences grounded on the predicted final goal to ensure on-task consistency (Section~\ref{backward}).
The subgoal sequences can effectively capture task progression and provide flexibility for downstream policy learning to leverage envisioned subgoals at varying distances.
To facilitate efficient policy training, we introduce a subgoal fusion technique that non-trivially compresses subgoal information and adaptively determines how to best utilize subgoals across different distances (Section~\ref{policy}).




\subsection{Grounding Task Objective as Latent Goals}\label{ground}
Previous research suggests that visual instructions can complement language descriptions, significantly enhancing guidance performance in conditioned visuomotor policy learning~\citep{radford2021learning,shah2023mutex,li2024decisionnce,zhenginstruction,li2024robo}. This synergy proves to be crucial in long-horizon, multi-stage tasks, where language descriptions often reduce to task identifiers due to their limited semantic information. 
In contrast, latent visual representations provide richer information about task progression, with latent visual goals offering precise specifications of the desired final scenario.
However, while latent visual goals can be easily obtained through hindsight labeling during training, their test-time specification presents challenges~\citep{lynch2020language}: it inherently depends on the current scenario configurations—for example, in the task ``place the brown cup in front of the white cup'', the precise goal state variably depends on the initialized relative spatial locations of the two cups. Crucially, this relationship is not fixed: if the position of the white cup changes at test time, the semantic meaning of ``in front of'' must be re-evaluated, requiring corresponding adjustments to the target visual goal.
To address this, we learn a goal prediction model $f_g$ that estimates the latent visual goal $z_g$ from the current observation $I_t$ and language instruction $l$. Given a dataset of video segments $\mathcal{D}_z = \{\tau_i\}^N$, with latent visual state $z_t$ in $\tau_i = \{(z_t, a_t)\}^{H_i}$ and language feature $\phi_l$ encoded by some pre-trained language-grounded visual encoder $(z_t, \phi_l) = \Phi(I_t, l)$,
we optimize:
{
\setlength{\abovedisplayskip}{6pt}
\setlength{\belowdisplayskip}{6pt}
\begin{equation}
    \max_{f_g} \sum_{\tau \in \mathcal{D}_z} \sum_{1 \leq t \leq H} \mathbb{E}_{p(z_g,\phi_l|\tau)} \log f_g(z_g|z_t,\phi_l)\label{latent_goal}
\end{equation}
}
where $p(z_g,l|\tau)$ represents the conditional context distribution ($p(c|\tau)$)
of latent goals and language instructions derived from trajectory $\tau$. 
This approach enables dynamic goal specification while ensuring the planning process operates within the semantically rich latent visual space.

\begin{figure}[t]
    \hspace{-10pt} 
    \includegraphics[width=0.52\textwidth]{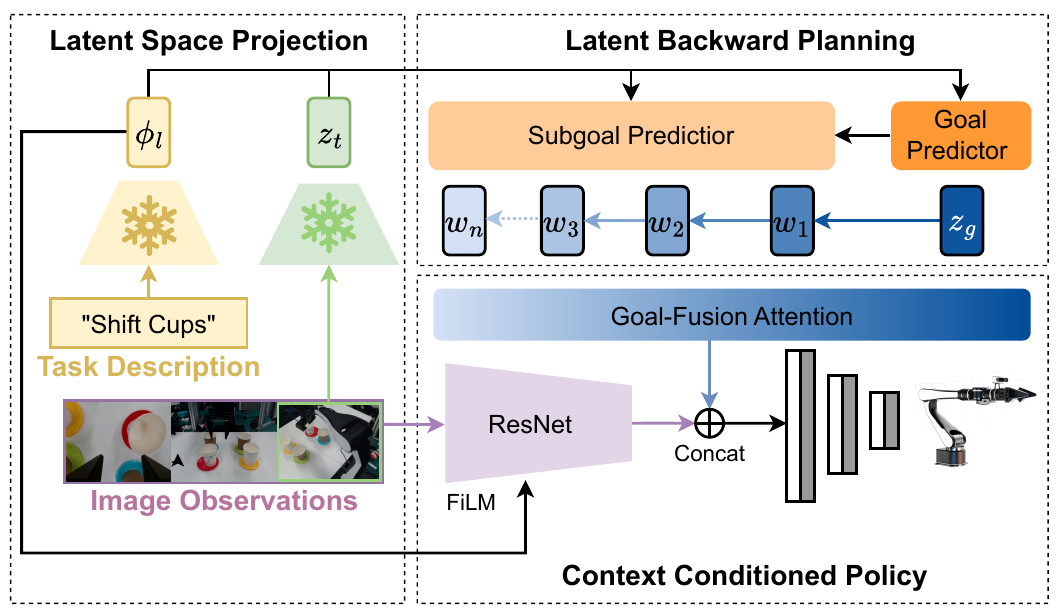}
    \caption{Overall framework architecture of LBP.}
    \label{fig:LBP-details}
    \vspace{-0.3cm}
\end{figure}

\subsection{Predicting Subgoals with a Backward Scheme}\label{backward}
While the final visual goal specifies the condition of task completion, it provides limited guidance about task progression—the sequence of states required to achieve the ultimate objectives. To better capture long-horizon task progression, we predict intermediate subgoals. 
However, subgoal selection presents a fundamental dilemma~\citep{park2024hiql,levy2018hierarchical}: balancing the sufficiency of subgoals for task progression against the accuracy of their prediction. Sparse subgoal predictions fail to adequately reflect task progression, while long subgoal sequences are prone to compounding prediction errors that lead to off-task behaviors deviating from the intended task goals.
To address this, we begin by predicting the first subgoal $w_1$ from the current state $z_t$, final goal $z_g$, and language instruction $\phi_l$ in latent space, with the optimization objective:
\begin{equation}
    \max_{f_w^1} \sum_{\tau \in \mathcal{D}_z} \sum_{1 \leq t \leq H} \mathbb{E}_{p(w_1,z_g,\phi_l|\tau)} \log f_w^1(w_1|z_t,z_g,\phi_l)\label{first_subgoal}
\end{equation}
To ensure sufficient long-term information, we set the first subgoal relatively close to the final goal, which maintains better alignment with task objectives yet provides less immediate guidance for policy learning. To bridge this gap, we recursively predict intermediate subgoals closer to the current state. Specifically, each subsequent subgoal $w_i$ is predicted from the previous subgoal $w_{i-1}$, current state $z_t$, and instruction $\phi_l$, forming a backward chain from coarse to fine temporal resolutions. The optimization objective for predicting subgoal $w_i$ is given by:
\begin{equation}
    \max_{f_w^i} \sum_{\tau \in \mathcal{D}_z} \sum_{1 \leq t \leq H} \mathbb{E}_{p(w_i,w_{i-1},\phi_l|\tau)} \log f_w^i(w_i|z_t,w_{i-1},\phi_l)\label{subgoal_seq}
\end{equation}

For convenience, let $\Gamma(w_i)$ denote the corresponding time step of subgoal $w_i$ in the trajectory. We can thus define a recursive planning coefficient $\lambda = \frac{\Gamma(w_i) - t}{\Gamma(w_{i-1}) - t}$, ($w_0=z_g$), to govern the recursive subgoal generation for $i=1,2,\cdots$, which represents the ratio of the temporal distance between the predicted subgoal and the current state $z_t$ relative to the distance between the previous-level subgoal $w_{i-1}$ and the current state $z_t$.

By inspecting Eq.~(\ref{first_subgoal}) and (\ref{subgoal_seq}), we can observe that it is possible to use a single unified model $f_w$ for all different levels of subgoal predictors $f_w^i$, as they all share the same structure.
This unified model is expected to predict the intermediate subgoal $z_\lambda := z_{\lceil(1-\lambda)t + \lambda k\rceil}$ between any start latent state $z_t$ and final latent state $z_k$, where $1 \leq t < k$. 
The objective is given by:
\begin{equation}\small
\begin{aligned}
&\max_{f_w} \sum_{\tau \in \mathcal{D}_z} \sum_{1 \leq t < H} \mathbb{E}_{\substack{p(z_H,\phi_l|\tau), \\ p(\{z_{\lambda^i}\}_{i=1}^n|\tau)}}
\left[ 
\sum_{i=1}^n \log f_w(z_{\lambda^i}|z_t,z_{\lambda^{i-1}},\phi_l) 
\right] \\
& +\sum_{\tau \in \mathcal{D}_z} \sum_{1 \leq t < H} \mathbb{E}_{\substack{p(z_H,\phi_l|\tau), \\ p(\{z_{\lambda^i}\}_{i=1}^n|\tau)}}
\left[ 
\sum_{i=1}^n \log f_w(z_{\lambda^i}|z_t,\hat{z}_{\lambda^{i-1}},\phi_l) 
\right]
\end{aligned}\label{interpolant_subgoal}
\end{equation}
where $z_{\lambda^i}:=z_{\lceil(1-\lambda^i)t+\lambda^i H\rceil}\subset\tau,\ i\in[1,\cdots,n]$ denotes the ground truth subgoal and $\hat{z}_{\lambda^i}$ denotes its predicted counterpart by $f_w$.
The first term in Eq.~(\ref{interpolant_subgoal}) fits subgoal prediction with the ground truths $z_{\lambda^i}$ in $\tau$, capturing the actual task progression.
The second term optimizes the subgoal predictor $f_w$ given 
its own previous predictions $\hat{z}_{\lambda^{i-1}}$ as inputs, ensuring the consistency of the recursive prediction of $f_w$ at test-time.
This recursive mechanism will suffer much less compounding error as the $\lambda$-recursion effectively reduces the planning steps, and the training of $f_w$ incorporates supervision of groundtruths in every recursion level.

As illustrated in Figure~\ref{fig:intro}, this backward planning scheme generates asymmetric coarse-to-fine grained latent subgoal sequences spanning the entire task horizon, offering three key advantages over conventional methods: (1) comprehensive task progression information in subgoal sequences, providing rich and flexible guidance for policy learning; (2) improved prediction consistency with task objectives in a backward manner, reducing error accumulation compared to forward planning; and (3) computational efficiency by adopting recursion, avoiding the need for fine-grained frame-by-frame prediction.

\subsection{Learning Context Conditioned Policy}\label{policy}
The generated subgoal sequence provides rich contextual information for policy learning. Given the complete context set $c = \{w_n, \dots, w_1, z_g, \phi_l\}\in\mathbb{R}^{(n+2)\times N_z}$ derived from dataset $\mathcal{D}_z$, we optimize the conditioned policy through:
\begin{equation}
    \max_\pi \sum_{\tau \in \mathcal{D}_z} \sum_{1 \leq t \leq H} \mathbb{E}_{c\sim p(c|\tau)} \log \pi_\theta(a_t|z_t,c)\label{policy_learning}
\end{equation}
However, even in latent space, the aggregated context dimensions can burden policy learning. Moreover, the policy should adaptively leverage (sub)goal information rather than treating all predictions equally, as different task execution stages require varied focus between short-term and long-term guidance. For instance, tasks requiring large movements intuitively benefit more from distant subgoals to prevent actions that hinder future progress, while precision-oriented tasks require stronger emphasis on nearby subgoals.

To address these challenges, we introduce a goal-fusion module with a Perceiver-style cross-attention~\citep{jaegle2021perceiver} that performs both correlation discovery and dimensionality reduction. Specifically, the contexts $c$ are queried by a trainable latent vector of size $D_z$: $z\in\mathbb{R}^{1\times N_z}$, which outputs the context embeddings $z_c$. This design compresses all contextual tokens $c$ into a lower-dimensional token $z_c$ while enabling the adaptive extraction of the most relevant context information.
This enables dynamic balancing of short- and long-term guidance throughout task execution, maximally leveraging the flexibility of predictions at varying distances and granularities.

\subsection{Practical Algorithm}


In the training phase, we learn a final goal predictor $f_g$ with Eq.~(\ref{latent_goal}), a unified subgoal predictor $f_w$ with Eq.~(\ref{interpolant_subgoal}), and a conditioned policy $\pi$ with Eq.~(\ref{policy_learning}). 
At each step $t$ at test time, LBP processes the current observation $I_t$ and language instruction $l$ into latent state $z_t$ and language feature $\phi_l$, and then generates latent (sub)goal plans $\{w_n, \dots, w_2, w_1, z_g\}$ by $f_g$ and $f_w$. Then we use the contexts $c$ including predicted goal plans and language features $\phi_l$ to condition the policy $\pi(a_t|I_t,c)$ for action extraction.


The detailed architecture of our model is present in Figure~\ref{fig:LBP-details}. We implement the goal predictor $f_g$ and the subgoal predictor $f_w$ using two-layer MLPs and employ a cross-attention block for the goal-fusion module.
Compared to recent planning-based methods that rely on complex pixel-level generative models, LBP demonstrates significant efficiency. For the low-level policy, we use a shared ResNet-34~\cite{he2016deep} as the backbone to extract visual features from different camera views, where the language embeddings are injected via FiLM conditioning layers~\cite{Film_perez2018film}. The current visual features encoded by ResNet are then integrated with the contexts to generate actions. The policy is optimized with diffusion loss to model complex distributions~\citep{chi2023diffusionpolicy}, with the denoising step fixed at 25. More details are provided in Appendix~\ref{app:implementations}.

\begin{table*}[!htp]
\centering
\vspace{-0.2cm}
\caption{\textbf{LIBERO-LONG results.} For each task, we present the average performance of top-3 checkpoints. 
The metric ``Avg. Success'' measures the average success rate across 10 tasks.
\model outperforms baselines with higher Avg. Success and better results on most tasks. The best results are \textbf{bolded}. LIBERO-LONG tasks include: \footnotesize(1) put soup and sauce in basket; (2) put box and butter in basket; (3) turn on stove and put pot; (4) put bowl in drawer and close it; (5) put mugs on left and right plates; (6) pick book and place it in back; (7) put mug on plate and put pudding to right; (8) put soup and box in basket; (9) put both pots on stove; (10) put mug in microwave and close it.}

\label{libero-long-main}
\vspace{3mm}
\scalebox{0.97}{\input{icml2024/tables/libero-long-main.tex}}
\vspace{-1cm}
\end{table*}

\section{Experiments}




\subsection{Experimental Setup}
Using previously validated evaluation recipes for embodied AI~\cite{blackzero_SuSIE,kim2024openvla,tian2024predictive_SEER,zheng2025universalactionsenhancedembodied}, we assess the performance of the proposed LBP.
Specifically, we assess LBP on both the LIBERO-LONG simulation benchmark and a real-robot environment with long-horizon, multi-stage tasks. 
For all methods involving subgoal prediction, the planning process is solely applied to the third-person view in the LIBERO-LONG experiments and the top view in the real-world experiments.
Unless otherwise stated, we adopt a three-step planning scheme (predicting a final goal and two intermediate subgoals) of LBP and set the planning coefficient $\lambda = 0.5$. Ablation studies on key framework designs and different choices of $\lambda$ are provided in Section~\ref{Ablations}.


\vspace{-2mm}
\paragraph{LIBERO-LONG experiments.} LIBERO-LONG~\cite{liu2024libero} consists of 10 distinct long-horizon robotic manipulation tasks that require diverse skills such as picking up objects, turning on a stove, and closing a microwave. These tasks involve multi-stage decision-making and span a variety of scenarios, making them particularly challenging. All models are trained on 50 unique expert demonstrations for each task. More details of LIBERO-LONG benchmark are provided in Appendix~\ref{app:libero-benchmark-details}.


\begin{figure*}[ht]
    \centering
    \includegraphics[width=1.0\linewidth]{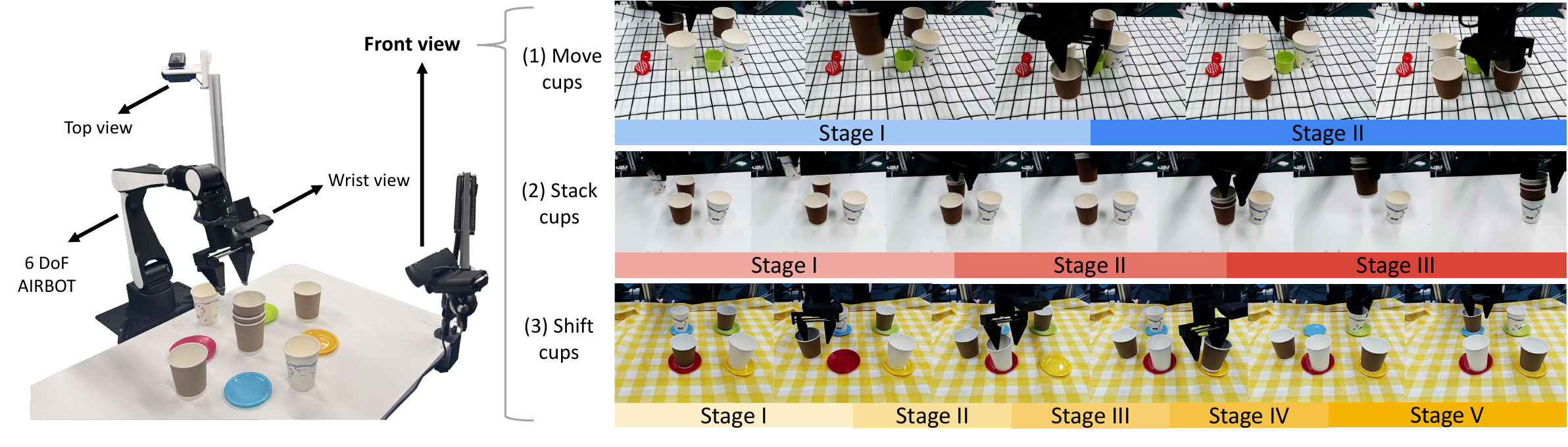}
    \caption{\textbf{Left:} the entire desktop environment setups of real-world experiments contains a 6 DoF AIRBOT arm and three Logitech C922PRO cameras with different views; \textbf{Right:} (1) \textit{Move cups:} move both brown cups in front of the white ones; (2) \textit{Stack cups:} stack all paper cups together; (3) \textit{Shift cups:} shift all the paper cups to another plate, in a clockwise direction. }
    \label{fig:setup}
    \vspace{0.2cm}
    \centering
    \includegraphics[width=1.0\linewidth]{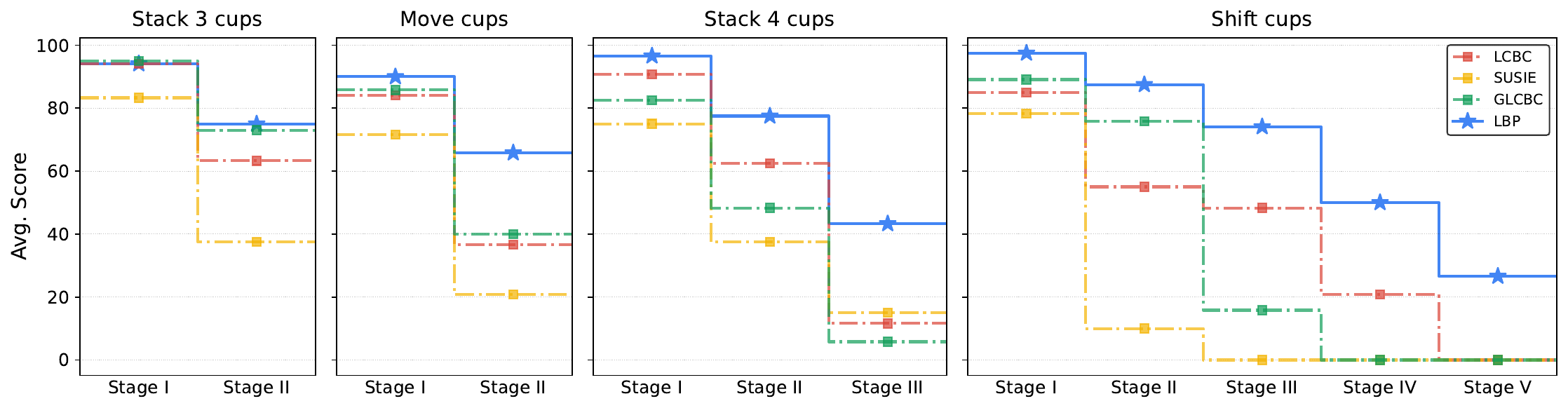}
    \vspace{-0.5cm}
    \caption{\textbf{Real-world main results}. We evaluate LCBC, GLCBC, SuSIE and LBP in aforementioned 4 tasks. The metric "Avg. Score" measures the average score for each stage. We observe that while LBP slightly outperforms other strong baselines at the early stages, LBP wins by a fairly large margin at the final stages of all tasks. This shows LBP significantly excels in handling long-horizon tasks.}
    \label{fig:Airbot-results}
    \vspace{-0.5cm}
\end{figure*}

\vspace{-2mm}
\paragraph{Real-world experiments.}
To investigate the effectiveness of LBP in real world, we specifically design four long-horizon tasks: \textit{Stack 3 cups}, \textit{Move cups}, \textit{Stack 4 cups} and \textit{Shift cups}. Each task is decomposed into multiple sequential stages, as illustrated in Figure~\ref{fig:setup}, requiring the robot to perform fundamental pick-and-place operations. 
These tasks establish a critical dependency where progress in subsequent stages is contingent on successful execution of preceding ones.
We assess task performance using a stage-based scoring system with discrete values {\{0, 25, 50, 75, 100\}} for each stage, where each score corresponds to the completion progress of the current stage.
A stage is assigned 100 only upon successful completion of the entire stage. 
All experimental evaluations are conducted with a 6 DoF AIRBOT robotic arm, together with three different views provided by Logitech C922PRO cameras. The overall environmental setups and task illustrations are shown in Figure~\ref{fig:setup}. All models are trained using 200 expert demonstrations for the task \textit{Move cups} and \textit{Shift cups}, and a total of 200 expert demonstrations for \textit{Stack 3 cups} and \textit{Stack 4 cups}. More details of experimental setups can refer to Appendix~\ref{app:real-benchmark-details}.


\paragraph{Baselines.}

For the LIBERO-LONG benchmark, we implement the multi-task policy MTACT~\citep{zhao2023learning_MTACT}, the general image-based pre-trained policy MVP ~\cite{xiao2022masked}, the interaction-oriented representation learning method MPI ~\cite{zeng2024learning}, large-scale pretrained vision-language-action policy OpenVLA ~\cite{kim2024openvla}, an image-editing based subgoal planner SuSIE~\cite{blackzero_SuSIE}, and the end-to-end predictive inverse dynamics model Seer ~\cite{tian2024predictive_SEER}. For real-world experiments, we compare LBP with SuSIE, one of the most competitive methods against LBP in the LIBERO-LONG benchmark. Also, we deploy vanilla Language Conditioned Behavior Cloning (LCBC) and Goal-and-Language Conditioned Behavior Cloning (GLCBC) for comprehensive comparison. 

\paragraph{Metrics for long-horizon multi-stage tasks.} 
Following Seer~\cite{tian2024predictive_SEER}, we evaluate model performance on the LIBERO-LONG benchmark by averaging the results of the top three checkpoints, each evaluated over 10 rollouts per task. For real-world experiments, we evaluate the last three checkpoints, with each checkpoint being tested across 10 rollouts per task to provide an average score at each stage, offering a thorough evaluation of long-horizon capabilities. Since each task consists of multiple stages, we design a fine-grained scoring system to evaluate performance at the stage level. Specifically, a score of \textbf{25} is awarded when the robot shows clear intent to approach the correct target object. If the target is successfully picked up, the score increases to \textbf{50}. Carrying the object toward the correct destination yields \textbf{75} points, and placing it successfully at the desired location results in a full score of \textbf{100}. For multi-stage tasks, we enforce a strict rule: the robot can proceed to the next stage only if the current stage achieves a full \textbf{100} points. This rule makes our real-robot experiments a rigorous benchmark for evaluating long-horizon capability.


\subsection{Main Results}
\paragraph{Simulation Experiment Results.}  
Table~\ref{tab:main-libero} presents the quantitative comparison on the LIBERO-LONG benchmark. LBP outperforms all baselines, achieving higher success rates across the majority of tasks. Specifically, LBP attains an average success rate of 85.0\% in SigLIP~\citep{zhai2023sigmoid_SigLIP} latent space and 88.6\% in DecisionNCE~\citep{li2024decisionnce} latent space, demonstrating its flexibility in leveraging different latent representations.  
Compared to SuSIE and Seer, which rely on heavy generative models for high-level planning, LBP demonstrates that lightweight MLPs can achieve comparable or even better performance in long-horizon tasks.
This improvement stems from the backward planning paradigm adopted in LBP, which maintains long-horizon consistency by recursively generating subgoals that preserve alignment with the final objective, ultimately enhancing both overall performance and computational efficiency on long-horizon multi-stage tasks.


\paragraph{Real-world experiment results.} 
In Figure~\ref{fig:Airbot-results}, we present the quantitative comparison on the real-world AIRBOT tasks. 
LBP consistently achieves the best performance at each stage in long-horizon tasks. Notably, in the early stages, the performance gap between different methods is relatively small. However, as the task progresses, other methods struggle due to insufficient and inconsistent guidance, leading to failures in later stages, whereas LBP maintains strong performance throughout.
The results also show that GLCBC sometimes initially outperforms LCBC by incorporating additional visual goal features but suffers a sharp decline in later stages. 
This drop is likely due to misalignment between the given final goals and current states, which misguides the policy in long-horizon tasks, highlighting the importance of dynamically predicting the final latent goal in LBP.
Additionally, we observe that SuSIE often generates hallucinated and incorrect subgoal images that confuse the low-level policy.
While this issue may be less pronounced in relatively deterministic simulation environments, it significantly impacts performance in real-world settings with inherently complex disturbances and stochasticity. 
In contrast, LBP enables easy prediction and efficient planning in latent space with its backward philosophy.


\begin{figure*}[t]
    \centering
    \includegraphics[width=1.0\linewidth]{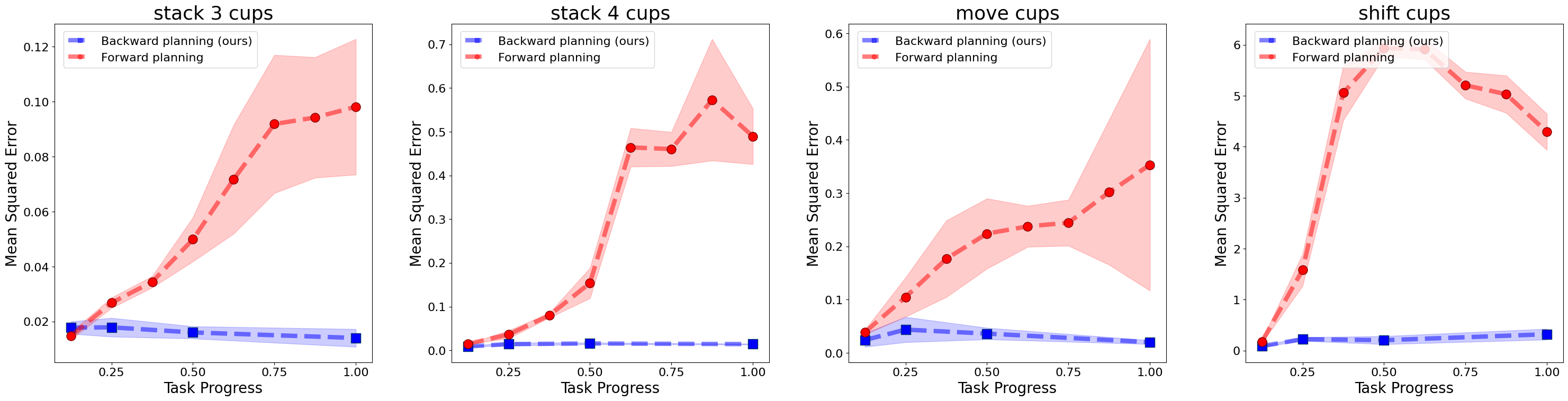}
    \vspace{-0.75cm}
    \caption{Mean Squared Errors (MSE) between predicted subgoals and corresponding ground truths under forward and backward paradigm.}
    \label{fig:backward-forward}
    \vspace{-0.5cm}
\end{figure*}

\paragraph{Comparison to the forward paradigm.} 

To evaluate the effectiveness of the backward planning paradigm, we compare it with the conventional forward planning, with both using the same hyperparameter configurations to ensure a fair comparison. Specifically, the forward planner predicts the subgoal 10 steps ahead at each iteration, autoregressively generating the entire subgoal sequence. We randomly sample 3,000 data points representing the current state from our real-robot datasets and compute the mean squared error (MSE) between the predicted subgoals and their corresponding ground truths. The results are visualized in Figure~\ref{fig:backward-forward}, with normalized task progress shown on the x-axis. 

We observe that the compounding errors of forward planning increase rapidly across all tasks. In particular, for the most challenging task, \textit{Shift Cups}, the prediction error becomes unacceptably large when predicting distant subgoals. This issue is further exacerbated in approaches that attempt to predict continuous future image frames, where compounding errors can be even more severe. In contrast, our backward planning method maintains consistently low error across the entire planning horizon. These results highlight the advantages of our approach, which enables both efficient and accurate subgoal prediction.

\subsection{Ablation Studies} \label{Ablations}


In this section, we conduct ablation studies to evaluate the impact of different design choices of LBP on long-horizon performance. All models adopt DecisionNCE latent space and are tested on LIBERO-LONG.

\begin{table}[t]
\centering
\caption{Ablations on different hyperparameter choices of LBP on LIBERO-LONG.}
\vspace{0.25cm}
\scalebox{1}{\input{icml2024/tables/libero-ablation-LBP.tex}}
\vspace{-1.25cm}
\end{table}

\paragraph{Ablation on key hyperparameters.} We perform an ablation study on two key hyperparameters: planning steps and the recursive planning coefficient $\lambda$ in Table~\ref{libero-ablation-LBP}. We test different numbers of planning steps, where more steps correspond to predicting more subgoals. Additionally, we vary the planning coefficient $\lambda$, which controls the temporal sparsity of the subgoal sequence—larger values result in more densely packed subgoals, closer to the final goal.
The main findings are:
(1) Without grounding the task objective in the latent visual goal \( z_g \), the approach reduces to LCBC, achieving an average success rate of 77.3\%. When \( z_g \) is provided as additional context, the variant shows a significant improvement of 6.0\%, demonstrating the effectiveness of leveraging the visual goal.
(2) Adding subgoals $w$ as additional contexts leads to an obvious performance improvement since it provides downstream policy with more about the future, but it is unnecessary to predict a large number of subgoals to achieve optimal results. This reflects the efficiency of our approach—unlike many planning methods that rely on generating numerous continuous waypoints, our method achieves high performance with fewer subgoals. This advantage likely arises from the backward planning philosophy of LBP, where subgoals are predicted recursively in reverse from the final goal, providing efficient yet relevant planning information closely aligned with task progression.
(3) We test \( \lambda  \) with 0.5 and 0.75, observing that LBP is robust and relatively insensitive to this hyperparameter choice.


\begin{table}[t]
\centering
\caption{Ablations on key design components of LBP on LIBERO-LONG.}
\vspace{0.5cm}
\scalebox{1}{\input{icml2024/tables/libero-ablation-component.tex}}
\vspace{-1.3cm}
\end{table}

\paragraph{Ablation on key architectural components.} 
We ablate the impact of the LBP planner and goal-fusion strategy, with results presented in Table~\ref{libero-ablation-component}. Removing the planner and relying solely on the low-level policy reduces the model to LCBC, resulting in a 11.3\% performance drop, underscoring the necessity of subgoals predicted by LBP.
Besides, replacing our goal-fusion strategy with simple average pooling causes a 9.6\% decline in performance, showing that naively compressing subgoals across different horizons undermines the low-level policy. This highlights the role of our goal-fusion strategy in adaptively leveraging subgoals at different distances in a way that effectively enhances planning performance.

\section{Conclusion and Future Direction}
We present LBP, a novel and efficient robotic planning framework that features backward planning in the latent space to break the critical trilemma among planning efficiency, long-horizon temporal consistency, and prediction accuracy. By leveraging visual latent space for planning, LBP achieves computational efficiency while maintaining sufficient information to capture task progression.
Moreover, by adopting the recursive ``coarse-to-fine" backward prediction paradigm, LBP fundamentally mitigates the compounding prediction errors inherent in traditional forward planning approaches, particularly addressing the challenges of off-task prediction in long-horizon scenarios. Extensive evaluations across diverse simulated and real-world environments, including complex long-horizon and multi-stage robotic tasks, consistently demonstrate LBP's superior performance and robustness. One promising research direction is the integration of advanced subgoal selection mechanisms, such as key-frame detection methods, to enhance the identification of informative subgoals. Another is the incorporation of more sophisticated robotic encoders to construct better-structured latent spaces for more efficient and accurate planning.

\section*{Acknowledgement}
This work is supported by funding from Wuxi Research Institute of Applied Technologies, Tsinghua University under Grant 20242001120, Beijing Academy of Artificial Intelligence (BAAI). 

\section*{Impact Statement}
This paper presents work whose goal is to advance the field of Machine Learning. There might be some potential societal consequences of our work, none which we feel must be specifically highlighted here.




\nocite{langley00}

\bibliography{example_paper}
\bibliographystyle{icml2025}

\newpage
\appendix
\onecolumn
\section{Implementation Details}\label{app:implementations}
\paragraph{LBP (Ours).} 
For high-level planner, we implement the goal predictor $f_g$ and the subgoal predictor $f_w$ using two-layer MLPs and employ two cross-attention blocks to realize the goal-fusion attention model. We employ DecisionNCE~\cite{li2024decisionnce} and SigLIP~\cite{zhai2023sigmoid_SigLIP} as frozen encoders to project language instructions and images to latent space. 

For the low-level policy, we use a shared ResNet-34~\cite{he2016deep} as the backbone to extract visual features from all camera view images, where the language embeddings are injected via FiLM conditioning layers~\cite{perez2018film}. The visual features, goal-fused feature, and current proprioception are then concatenated and fed into a residual MLP~\cite{ResMLP_hansen2023idql} to generate actions. The policy is optimized with diffusion loss to model complex distributions~\cite{chi2023diffusionpolicy}, with the denoising step fixed at 25. We employ action chunking with a fixed length of 6 and utilize an exponential weighting scheme to ensemble overlapping action sequences, following~\cite{zhao2023learning_MTACT}.

For training the high-level planner, we use a batch size of 64 and train for 100k steps with the AdamW optimizer. For the low-level policy on LIBERO-LONG, we set the batch size to 64 and train for 200k steps. In the case of the low-level policy for real-world robot experiments, we increase the batch size to 128 and train for 400k steps.

\paragraph{SuSIE~\citep{blackzero_SuSIE}.}
The high-level image-editing diffusion model is trained on video data using four A6000 GPUs. We utilize the official codebase with minimal modifications, altering only the datasets. For simulation experiments, we fine-tune the released SuSIE checkpoint on the LIBERO dataset. 
In the real-robot experiments, we fine-tune the checkpoint exclusively with collected data on Airbot.
Regarding the low-level policy, we adopt the same model architecture with the policy model of LBP, while adopting a channel-wise concatenation of subgoal images and current images as inputs. Additionally, language instructions are removed to maintain consistency with the downstream training in SuSIE.

\paragraph{LCBC.} We implement it by directly removing our high-level planner from the architecture of LBP. The language is projected to the latent space by CLIP~\citep{radford2021learning}, and images are projected by a ResNet-34~\citep{he2016deep}, then the semantic features are captured by a FiLM~\citep{perez2018film} module. The low-level policy takes in these semantics, together with current proprioception, and then outputs a predicted diffusion noise.

\paragraph{GLCBC.} The only difference between GLCBC and LCBC is the part before entering the FiLM module: we choose a predefined image as the final goal and then project it to a latent space with DecisionNCE image encoder~\citep{li2024decisionnce}. We concatenate the language embeddings with the final goal image embeddings, then incorporate them as inputs into the FiLM module. 

\paragraph{Others.} For LIBERO-LONG benchmark, since our experimental settings and evaluation metrics are the same with Seer, we obtain the scores of MTACT, MVP, MPI, OpenVLA and Seer from the original paper~\cite{tian2024predictive_SEER}.

\begin{table*}[h]
\centering
\caption{Language instructions and average lengths of LIBERO-LONG.}  
\vspace{3mm}
\scalebox{0.8}{\input{icml2024/tables/libero-dataset}}
\vspace{-1cm}
\end{table*}

\section{LIBERO-LONG Benchmark Details}\label{app:libero-benchmark-details}
\label{libero_detail}
We follow ~\citep{kim2024openvla} to re-render the images at a resolution of 256×256. The detailed language instructions and average demonstration lengths for each task of LIBERO-LONG is shown in Table~\ref{libero_dataset}.

\section{Real-Robot Experiment Details} \label{app:real-benchmark-details}
\label{airbot_detail}

\begin{table*}[htbp]
\centering
\caption{Dataset settings of real robot experiments.}  
\scalebox{0.8}{\input{icml2024/tables/real-robot-dataset}}
\vspace{-1cm}
\end{table*}

We collect 200 expert demonstrations each for tasks \textit{Move cups}, \textit{Stack 3 cups}, \textit{Stack 4 cups} and \textit{Shift cups}. To enhance the robustness of model trained on this dataset, we manually add some augmentation techniques (Table~\ref{airbot_dataset}), including \textit{Distractor augmentation}, \textit{Target augmentation}, \textit{Background augmentation} and \textit{View augmentation}. Examples from the side view can be seen at Table~\ref{example}. \textbf{View augmentation} always exsits because the side view camera is not a fixed-position view. \textbf{Distractor augmentation} means placing various unrelated distractor objects on the table. \textbf{Target augmentation} refers to replacing the target objects with different ones of this type, i.e., replacing paper cups with cups of different materials. \textbf{Background augmentation} is placing tablecloths with various colors above the clean white table.


\section{Additional Results}\label{app:results}
\paragraph{Numerical results.} Except for the visualized results of real-robot experiments in Figure~\ref{fig:Airbot-results}, we also present the numerical results of each stage for each task in Table~\ref{tab:stack_3_cups}-\ref{tab:shift_cups}.

\paragraph{Generalization experiment.} We test LBP on the longest real-world task (\textit{Shift Cups}) with different backgrounds and distractor objects, where we find that LBP maintains robust performance in these complex scenarios, still outperforming the strongest baseline LCBC in base setting. The numerical results are present in Table~\ref{tab:gen}.

\paragraph{Comparison to the parallel planning paradigm.} 

\begin{figure*}[h]
    \centering
    \includegraphics[width=1.0\linewidth]{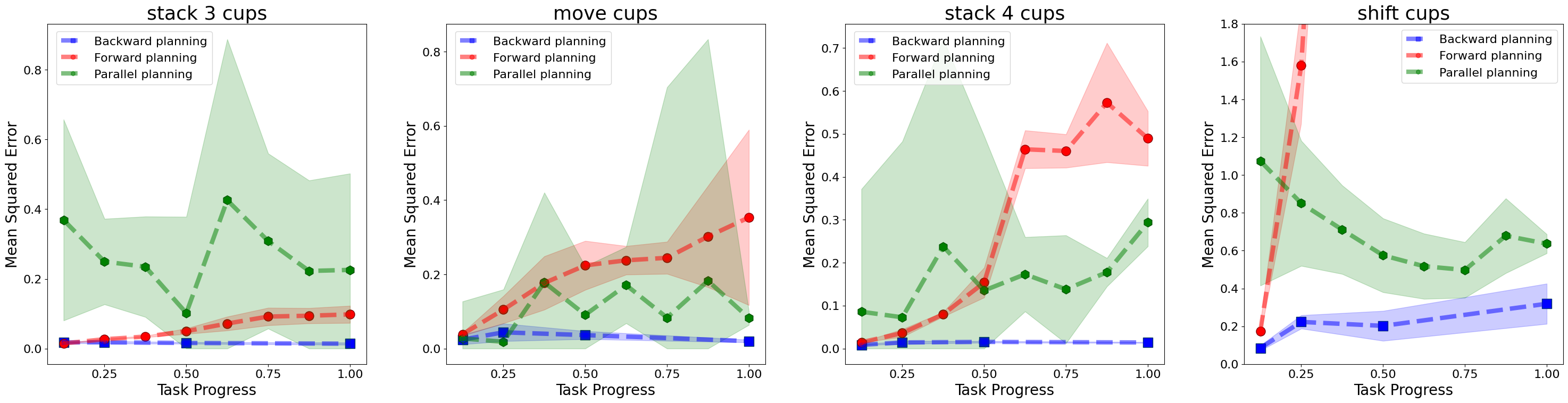}
    \caption{Mean Squared Errors (MSE) between predicted subgoals and corresponding ground truths in parallel, forward and backward planning.}
    \label{fig:backward-parallel-forward}
    \vspace{-0.5cm}
\end{figure*}

To evaluate the effectiveness of the backward planning paradigm, we further compare it with the parallel planning paradigm~\cite{janner2022planning,ajay2022conditional}, with both using the same hyperparameter configurations with LBP to ensure a fair comparison. Specifically, the parallel planner predicts all subgoals simultaneously in a single batch at each execution step. We randomly sample 3,000 data points representing the current state from our real-robot datasets and compute the mean squared error (MSE) between the predicted subgoals and their corresponding ground truths. The results are visualized in Figure~\ref{fig:backward-parallel-forward}, with normalized task progress shown on the x-axis. 

We observe that although parallel planning avoids error accumulation by predicting all subgoals simultaneously, it suffers from consistently inaccurate predictions across the entire planning horizon. This limitation can be attributed to the difficulty of the training objective, which requires simultaneous supervision of all subgoals. Such an approach demands greater model capacity and incurs significantly higher computational costs. In contrast, our backward planning method maintains consistently low prediction error across the entire planning horizon. These results highlight the advantages of our approach, which enables both efficient and accurate subgoal prediction.

\begin{table*}[htbp]
\centering
\caption{Demonstrations of augmentation techniques.}
\vspace{3mm}
\scalebox{0.8}{\input{icml2024/tables/augmentation_example}}
\vspace{-1cm}
\end{table*}

\begin{table*}[htbp]
\centering
\caption{Numerical results of task: \textbf{\textit{Stack 3 cups}}.}  
\vspace{3mm}
\scalebox{0.9}{\input{icml2024/tables/stack_3_cups}}
\vspace{-0.5cm}
\end{table*}

\begin{table*}[htbp]
\centering
\caption{Numerical results of task: \textbf{\textit{Move cups}}.} 
\vspace{3mm}
\scalebox{0.9}{\input{icml2024/tables/move_cups}}
\vspace{-0.5cm}
\end{table*}

\begin{table*}[htbp]
\centering
\caption{Numerical results of task: \textbf{\textit{Stack 4 cups}}.} 
\vspace{3mm}
\scalebox{0.9}{\input{icml2024/tables/stack_4_cups}}
\vspace{-0.5cm}
\end{table*}

\begin{table*}[htbp]
\centering
\caption{Numerical results of task: \textbf{\textit{Shift cups}}.} 
\vspace{3mm}
\scalebox{0.9}{\input{icml2024/tables/shift_cups}}
\vspace{-1cm}
\end{table*}

\begin{table*}[h]
\centering
\caption{Numerical results of generalization experiment on \textbf{\textit{Shift cups}}.} 
\vspace{3mm}
\scalebox{0.9}{\input{icml2024/tables/generalization}}
\vspace{-1cm}
\end{table*}

\end{document}

%% file: icml2024/tables/libero-long-main.tex
\normalsize
\renewcommand{\arraystretch}{1.2}
\begin{tabular}{x{60}x{25}x{25}x{25}x{25}x{25}x{25}x{25}x{25}x{25}x{25}x{45}}
\toprule
\resizebox{60pt}{!}{\diagbox{Method}{$\text{Task ID}$}} & 1 & 2 & 3 & 4 & 5 & 6 & 7 & 8 & 9 & 10 & Avg. Suc $\uparrow$ \\ \hline

MTACT & 0.00 & 50.0 & 75.0 & 85.0 & 20.0 & 75.0 & 0.00 & 30.0 & 10.0 & 65.0 & 41.0 \\
MVP & 78.3 & 90.0 & 80.0 & 88.3 & 46.6 & 63.3 & 45.0 & 83.3 & 60.0 & 46.6 & 68.2 \\
MPI & 86.6 & 86.6 & 96.6 & 95.0 & 83.3 & 83.3 & 56.6 & 66.6 & 40.0 & 78.3 & 77.3 \\
OpenVLA & 45.0 & 95.0 & 65.0 & 45.0 & 40.0 & 80.0 & 60.0 & 35.0 & 20.0 & 55.0 & 54.0  \\
Seer & 88.3 & 90.0 & 91.6 & 81.6 & \textbf{85.0} & 65.0 & 86.6& 80.0 & 51.6 & 66.6 & 78.6 \\ 
$\text{SuSIE}^{\dagger}$ & 83.3 & 63.3 & 96.6 & \textbf{100.0} & 83.3 & 83.3 & 83.3 & 39.9 & 53.3 & 76.6 & 76.3  \\
\rowcolor{gray!20}$\text{LBP}_{\textit{SigLIP}}$ &86.6 & \textbf{100.0} & 93.3 & \textbf{100.0} & 63.3 & 73.3 & 86.6 & 80.0 & \textbf{73.3} & 93.3  & 85.0  \\
\rowcolor{gray!20}$\text{LBP}_{\textit{DecisionNCE}}$ & \textbf{90.0}&\textbf{100.0}&\textbf{100.0}&\textbf{100.0}&76.6&\textbf{86.6}&\textbf{90.0}&\textbf{86.6}&60.0&\textbf{96.6}&\textbf{88.6}\\

\bottomrule 
\label{tab:main-libero}
\vspace{-2.5mm}

\parbox{\textwidth}{\small $\dagger$ \textit{Since the original SuSIE only supports single-view input, we incorporate a wrist view to reproduce it for fair comparison.}}
\vspace{3mm}
\end{tabular}

%% file: icml2024/tables/libero-ablation-LBP.tex



\normalsize
\renewcommand{\arraystretch}{1.2}
\begin{tabular}{x{50}x{80}x{55}}
\hline
$\lambda$ &  (Sub)Goals & Avg. Suc $\uparrow$ \\ \hline
 - & - & 77.3\\
 - & $z_g$ & 83.3 \\ \hline
0.5 & $z_g, w_1$ & 85.6 \\
0.5 & $z_g, w_1, w_2$ & \textbf{88.6} \\ 
0.5 & $z_g, w_1, w_2, w_3$ & 83.0 \\ \hline
0.75 & $z_g, w_1$ & 84.6 \\ 
0.75 & $z_g, w_1, w_2$ & 85.0 \\
0.75 & $z_g, w_1, w_2, w_3$ & 84.0 \\ \hline

\label{libero-ablation-LBP}
\end{tabular}

%% file: icml2024/tables/libero-ablation-component.tex
\normalsize
\renewcommand{\arraystretch}{1.2}
\begin{tabular}{x{70}x{70}x{45}}
\hline
 & Variant & Avg. Suc $\uparrow$ \\ \hline
\multirow{2}{*}{{\makecell[c]{Effectiveness of \\ the planner}}} & w/o planner & 77.3 \\ \cline{2-3}
& ours & \textbf{88.6} \\ \hline

\multirow{2}{*}{{\makecell[c]{The strategy \\ of goal-fusion}}} & average pooling & 79.0 \\ \cline{2-3}
& ours & \textbf{88.6} \\ \hline
\label{libero-ablation-component}
\end{tabular}

%% file: icml2024/tables/libero-dataset.tex
\normalsize
\renewcommand{\arraystretch}{1.5}
\begin{tabular}{>{\centering\arraybackslash}x{60}>{\centering\arraybackslash}x{100}>{\centering\arraybackslash}x{180}>{\centering\arraybackslash}x{100}}
\toprule
\makecell[c]{Task\\ID} & \makecell[c]{Task\\name} & \makecell[c]{Language\\instruction} & \makecell[c]{Average demonstration\\length (frames)} \\ \hline
\makecell[c]{1} & \makecell[c]{put soup and\\sauce in basket} & \makecell[c]{put both the alphabet soup\\and the tomato sauce in the basket} & \makecell[c]{294} \\ \hline
\makecell[c]{2} & \makecell[c]{put box and\\butter in basket} & \makecell[c]{put both the cream cheese\\box and the butter in the basket} & \makecell[c]{260} \\ \hline
\makecell[c]{3} & \makecell[c]{turn on stove\\and put pot} & \makecell[c]{turn on the stove and\\put the moka pot on it} & \makecell[c]{266} \\ \hline
\makecell[c]{4} & \makecell[c]{put bowl in\\drawer and close it} & \makecell[c]{put the black bowl in the bottom\\drawer of the cabinet and close it} & \makecell[c]{249} \\ \hline
\makecell[c]{5} & \makecell[c]{put mugs on\\left and right plates} & \makecell[c]{put the white mug on the\\left plate and put the yellow\\and white mug on the right plate} & \makecell[c]{258} \\ \hline
\makecell[c]{6} & \makecell[c]{pick book and\\place it in back} & \makecell[c]{pick up the book and place it in\\the back compartment of the caddy} & \makecell[c]{189} \\ \hline
\makecell[c]{7} & \makecell[c]{put mug on plate and\\put pudding to right} & \makecell[c]{put the white mug on\\the plate and put the chocolate\\pudding to the right of the plate} & \makecell[c]{255} \\ \hline
\makecell[c]{8} & \makecell[c]{put soup and\\box in basket} & \makecell[c]{put both the alphabet soup and\\the cream cheese box in the basket} & \makecell[c]{270} \\ \hline
\makecell[c]{9} & \makecell[c]{put both\\pots on stove} & \makecell[c]{put both moka\\pots on the stove} & \makecell[c]{416} \\ \hline
\makecell[c]{10} & \makecell[c]{put mug in\\microwave and close it} & \makecell[c]{put the yellow and white\\mug in the microwave and close it} & \makecell[c]{305} \\
\bottomrule 
\label{libero_dataset}
\vspace{-2.5mm}
\end{tabular}

%% file: icml2024/tables/real-robot-dataset.tex
\normalsize
\renewcommand{\arraystretch}{1.5}
\begin{tabular}{>{\centering\arraybackslash}x{70}>{\centering\arraybackslash}x{180}>{\centering\arraybackslash}x{70}>{\centering\arraybackslash}x{70}>{\centering\arraybackslash}x{70}>{\centering\arraybackslash}x{70}}
\toprule
\makecell[c]{Task\\name} & \makecell[c]{Language\\instruction} & \makecell[c]{w/ distractor\\augmentation?} & \makecell[c]{w/ view\\augmentation?} & \makecell[c]{w/ target\\augmentation?} & \makecell[c]{w/ background\\augmentation?} \\ \hline

\makecell[c]{Move cups} & \makecell[c]{first put the right brown cup in front\\ of the right white cup then put the left\\ brown cup in front of the left white cup} & \makecell[c]{\checkmark} & \makecell[c]{\checkmark} \\ \hline
\makecell[c]{Stack 3/4 cups} & \makecell[c]{stack the paper cups} & \makecell[c]{\checkmark} & \makecell[c]{\checkmark} \\ \hline
\makecell[c]{Shift cups} & \makecell[c]{move each cup to a new\\ position in a clockwise direction} & \makecell[c]{\checkmark} & \makecell[c]{\checkmark} & \makecell[c]{\checkmark} & \makecell[c]{\checkmark}\\

\bottomrule 
\label{airbot_dataset}
\vspace{-2.5mm}
\end{tabular}

%% file: icml2024/tables/augmentation_example.tex
\normalsize
\renewcommand{\arraystretch}{1.5}
\begin{tabular}{>{\centering\arraybackslash}x{110}>{\centering\arraybackslash}x{400}}
\toprule
\makecell[c]{Augmentation metrics} & \makecell[c]{Examples} \\ \hline

\makecell[c]{Distractor augmentation} &\makecell[c]{\raisebox{0pt}{\includegraphics[width=0.25\linewidth]{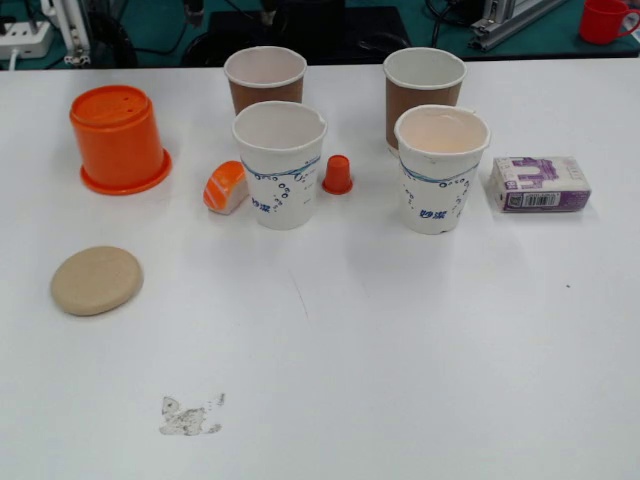}} \raisebox{0pt}{\includegraphics[width=0.25\linewidth]{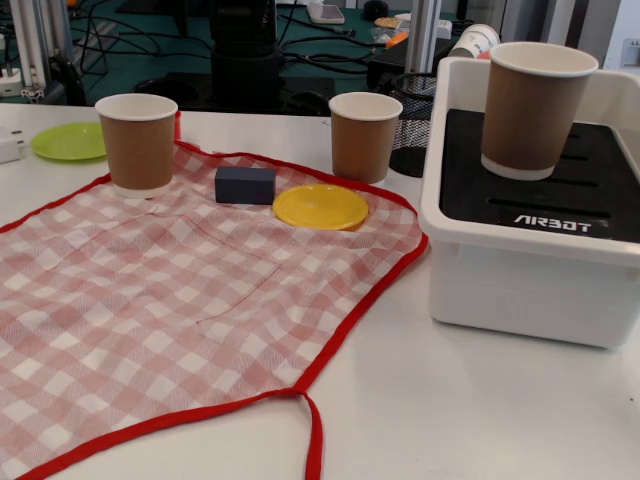}} 
\raisebox{0pt}{\includegraphics[width=0.25\linewidth]{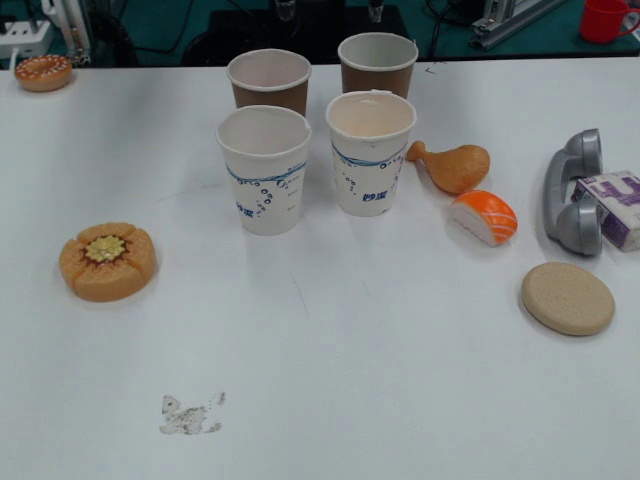}}} \\ \hline
\makecell[c]{Target augmentation} &\makecell[c]{\raisebox{0pt}{\includegraphics[width=0.25\linewidth]{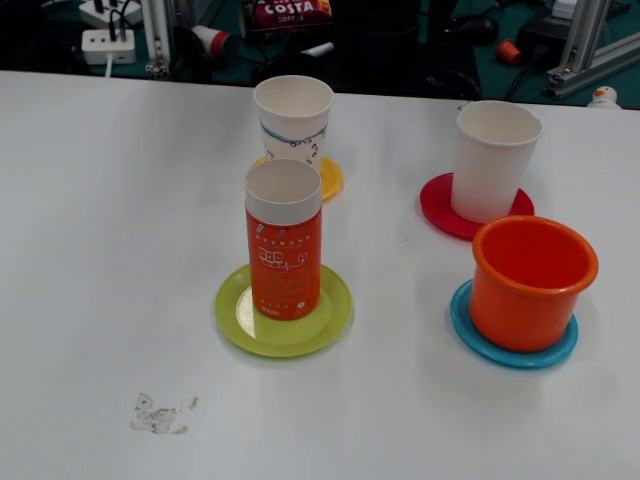}} \raisebox{0pt}{\includegraphics[width=0.25\linewidth]{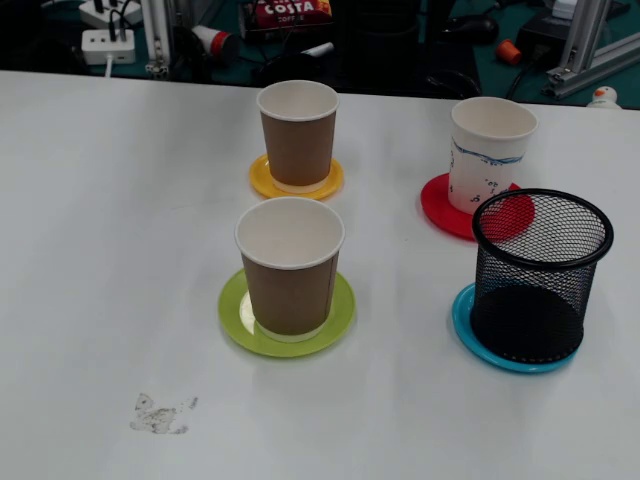}} 
\raisebox{0pt}{\includegraphics[width=0.25\linewidth]{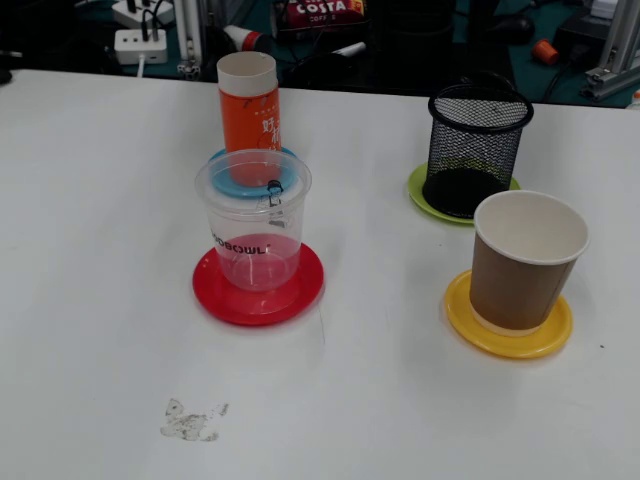}}} \\ \hline
\makecell[c]{Background augmentation} &\makecell[c]{\raisebox{0pt}{\includegraphics[width=0.25\linewidth]{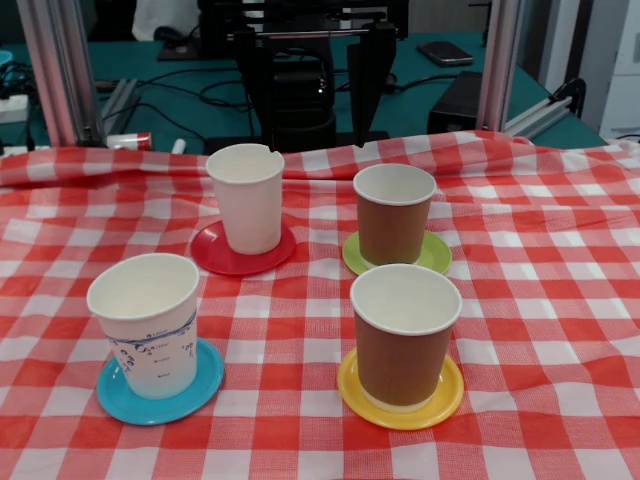}} \raisebox{0pt}{\includegraphics[width=0.25\linewidth]{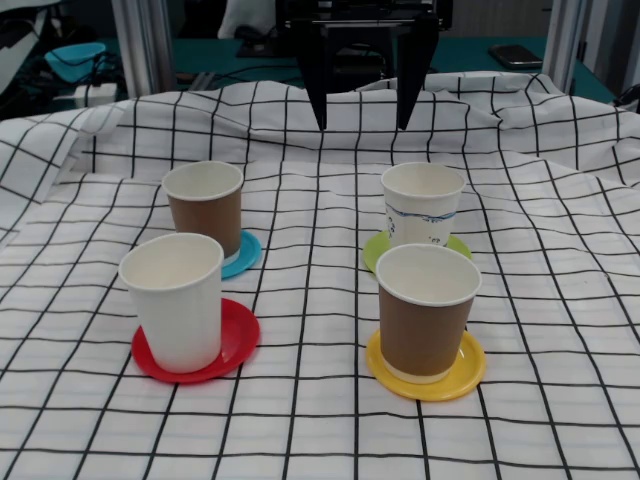}} 
\raisebox{0pt}{\includegraphics[width=0.25\linewidth]{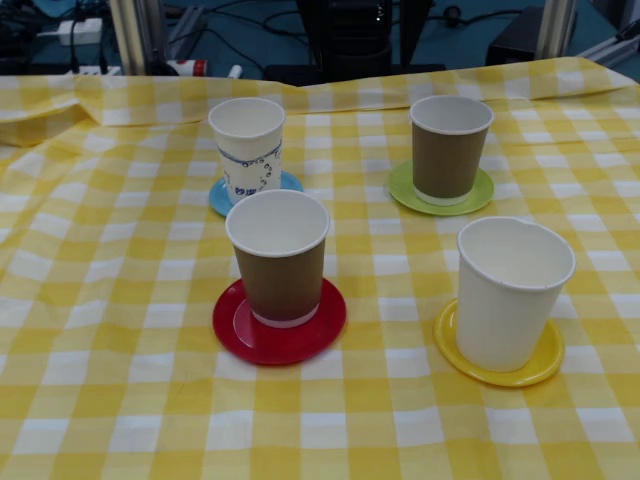}}} \\

\bottomrule 
\label{example}
\vspace{-2.5mm}
\end{tabular}

%% file: icml2024/tables/stack_3_cups.tex
\normalsize
\renewcommand{\arraystretch}{1.4}
\begin{tabular}{x{60}x{45}x{45}x{60}}
\toprule
Method & Stage I & Stage II & Avg. Score $\uparrow$ \\ \hline

LCBC & 94.1 & 63.3 & 78.7 \\
GLCBC & \textbf{95.0} & 74.1 & \textbf{84.6} \\
SuSIE & 83.3 & 37.5 & 60.4 \\
\rowcolor{gray!20}$\text{LBP}$ & 94.1 & \textbf{75.0} & \textbf{84.6} \\

\bottomrule 
\label{tab:stack_3_cups}
\vspace{-2.5mm}
\end{tabular}

%% file: icml2024/tables/move_cups.tex
\normalsize
\renewcommand{\arraystretch}{1.4}
\begin{tabular}{x{60}x{45}x{45}x{60}}
\toprule
Method & Stage I & Stage II & Avg. Score $\uparrow$ \\ \hline

LCBC & 84.1 & 36.6 & 60.4 \\
GLCBC & 85.8 & 40.0 & 62.9 \\
SuSIE & 71.6 & 20.8 & 46.2 \\
\rowcolor{gray!20}$\text{LBP}$ & \textbf{90.0} & \textbf{65.8} & \textbf{77.9} \\

\bottomrule 
\label{tab:move_cups}
\vspace{-2.5mm}
\end{tabular}

%% file: icml2024/tables/stack_4_cups.tex
\normalsize
\renewcommand{\arraystretch}{1.4}
\begin{tabular}{x{60}x{45}x{45}x{45}x{60}}
\toprule
Method & Stage I & Stage II & Stage III & Avg. Score $\uparrow$ \\ \hline

LCBC & 90.8 & 62.5 & 11.6 & 55.0 \\
GLCBC & 82.5 & 48.3 & 5.8 & 45.5 \\
SuSIE & 75.0 & 37.5 & 15.0 & 42.5\\
\rowcolor{gray!20}$\text{LBP}$ & \textbf{96.6} & \textbf{77.5}  & \textbf{43.3} & \textbf{72.5} \\

\bottomrule 
\label{tab:stack_4_cups}
\vspace{-2.5mm}
\end{tabular}

%% file: icml2024/tables/shift_cups.tex
\normalsize
\renewcommand{\arraystretch}{1.4}
\begin{tabular}{x{60}x{45}x{45}x{45}x{45}x{45}x{60}}
\toprule
Method & Stage I & Stage II & Stage III & Stage IV & Stage V & Avg. Score $\uparrow$ \\ \hline

LCBC & 85.0 & 55.0 & 48.3 & 20.8 & 0.0 & 41.8 \\
GLCBC & 89.1 & 75.8 & 15.8 & 0.0 & 0.0 & 36.1 \\
SuSIE & 78.3 & 10.0 & 0.0 & 0.0  & 0.0 & 17.7\\
\rowcolor{gray!20}$\text{LBP}$ & \textbf{97.5} & \textbf{87.5}  & \textbf{74.1} & \textbf{50.0} & \textbf{26.6} & \textbf{67.1} \\

\bottomrule 
\label{tab:shift_cups}
\vspace{-2.5mm}
\end{tabular}

%% file: icml2024/tables/generalization.tex
\normalsize
\renewcommand{\arraystretch}{1.4}
\begin{tabular}{x{120}x{45}x{45}x{45}x{45}x{45}x{60}}
\toprule
Method & Stage I & Stage II & Stage III & Stage IV & Stage V & Avg. Score $\uparrow$ \\ \hline

LCBC (Base setting)	& 85.0	& 55.0 & 48.3 & 20.8 & 0.0 & 41.8\\
LBP (Distracting objects) 	& 87.5 & 75.8 & 48.3 & 35.0 & 9.0 & 51.1\\
LBP (Different backgrounds)	& 91.6 & 84.1 & 55.8 & 37.5 & 13.3 & 56.4\\
LBP (Base setting)	& 97.5	& 87.5 & 74.1 & 50.0 & 26.6 & 67.1\\

\bottomrule 
\label{tab:gen}
\vspace{-2.5mm}
\end{tabular}